\documentclass[letterpaper]{article} 
\usepackage{aaai2027}  
\usepackage[hyphens]{url}  
\usepackage{graphicx} 
\usepackage{natbib}  
\usepackage{caption} 
\usepackage{algorithm}
\usepackage{algorithmic}
\usepackage{amsfonts}
\usepackage{multirow}
\usepackage{newfloat}
\usepackage{listings}
\usepackage{placeins}
\DeclareCaptionStyle{ruled}{labelfont=normalfont,labelsep=colon,strut=off} 
\floatstyle{ruled}
\newfloat{listing}{tb}{lst}{}
\floatname{listing}{Listing}

\usepackage{booktabs}
\usepackage{colortbl}

\usepackage{amsmath}

\title{OPRD: On-Policy Reminder Distillation for Long-Horizon Coding Agents}
\author{
    Yuechen Luo\textsuperscript{\rm 1,}\thanks{Equal contribution.},
    Zhiyi Lai\textsuperscript{\rm 2,*},
    Miao Ziqi\textsuperscript{\rm 3,*},
    Zihang Xu\textsuperscript{\rm 3},
    Jingxuan Xu\textsuperscript{\rm 4},\\
    Songwei Yu\textsuperscript{\rm 4},
    Wenqiang Zhu\textsuperscript{\rm 2},
    Weihao Li\textsuperscript{\rm 2},
    Junhao Fan\textsuperscript{\rm 1}, \\
    Haoyang Huang\textsuperscript{\rm 2}, 
    Yanan Wu\textsuperscript{\rm 2},
    Ken Deng\textsuperscript{\rm 2,}\thanks{Corresponding author.}
}
\affiliations{
    \textsuperscript{\rm 1}Tsinghua University,
    \textsuperscript{\rm 2}Independent Researcher\\
    \textsuperscript{\rm 3}Shanghai Artificial Intelligence Laboratory,
    \textsuperscript{\rm 4}National University of Singapore\\    \texttt{luo-yc24@mails.tsinghua.edu.cn}
}

\begin{document}

\maketitle

\begin{abstract}
    Reliable repository-level coding requires agents not only to generate correct code but also to preserve constraints and respond to tool feedback across long interactions. However, our diagnostic analysis shows that agents often fail to act on relevant information already available in context, while final Pass/Fail outcomes provide only sparse process supervision. We introduce \textbf{On-Policy Reminder Distillation (OPRD)}, which combines Multi-Scope Reminder supervision with outcome-based routing on reminder-free student rollouts. Successful rollouts provide autoregressive self-imitation targets, whereas failed rollouts receive Reminder-conditioned reverse-KL supervision from a teacher that re-evaluates the same student-generated prefixes. A Global Reminder converts final-test diagnostics into trajectory-level guidance, while Turn-local Reminders apply only to the next decision following the relevant environment observation. The teacher and student share a two-stage initialization from Coding SFT and Instruction-Following (IF) SFT. At inference, the student operates without Reminders. On \textsc{SWE-bench Verified}, \textsc{SWE-bench Pro}, and \textsc{SWE-bench Multilingual}, OPRD improves the benchmark-averaged task resolution rate by 11.3 points over the open-source Qwen3.5-4B base model. Under the same diagnostic protocol, it reduces the aggregate error-event count from 446 to 222, a 50.2\% reduction.
\end{abstract}

\section{Introduction}
\label{sec:introduction}

\begin{figure}[t]
    \centering
    \includegraphics[width=\columnwidth]{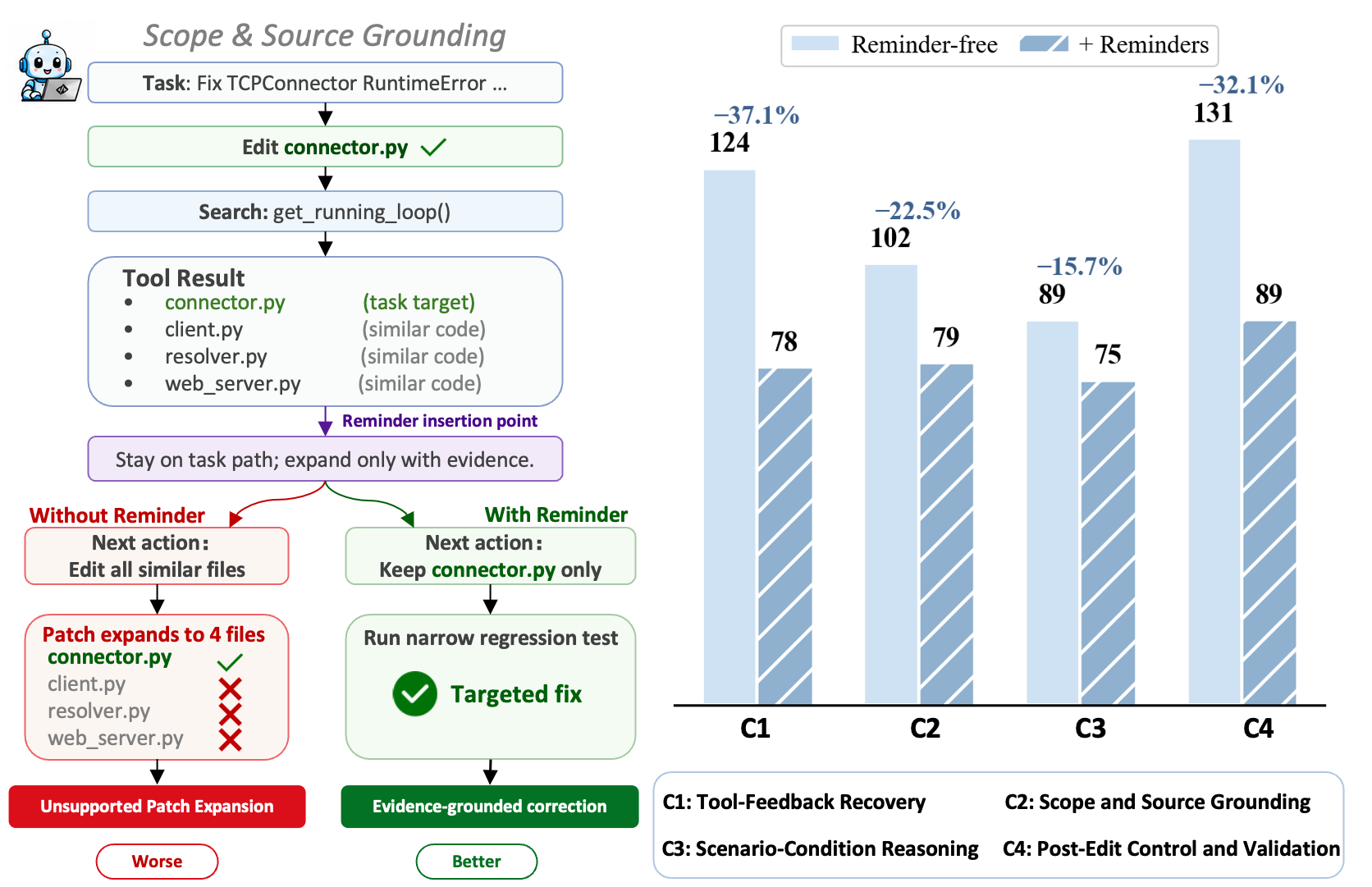}
    \caption{Diagnostic analysis of Qwen3.5-4B on 200 tasks from the \textsc{SWE-rebench} v2 validation set. Left: a \textit{Scope and Source Grounding} example in which a Turn-local Reminder prevents unsupported patch expansion. Right: aggregate error-event counts across four categories under Reminder-free and Reminder-augmented settings; percentages indicate relative reductions.}
    \label{fig:intro}
    \end{figure}

Large language models have advanced in reasoning~\citep{zhan2025katv1,deng2025hipo,xu2026stableadaptive,miao2026seeing} and repository-level software engineering~\citep{yang2025codeintelligencesurvey,jimenez2024swebench,yang2024sweagent,zhang2024autocoderover,ouyang2025repograph,xia2025agentless}. Reliable long-horizon coding nevertheless requires agents to preserve task constraints and adapt to tool and execution feedback~\citep{yao2023react,chen2024selfdebug}. We examine this challenge on 200 \textsc{SWE-rebench} V2 validation tasks~\citep{badertdinov2026swerebenchv2}, using two Qwen3.5-4B rollouts per task under identical inference settings~\citep{qwen2026qwen35}. In the reminder-free rollout, deterministic rules identify explicit tool and execution failures, while GPT-5.4 classifies remaining context-dependent events from the task and complete trajectory~\citep{openai2026gpt54}. Each event belongs to one of four categories in Table~\ref{tab:reminder-taxonomy-summary}: \textit{Tool-Feedback Recovery}, \textit{Scope and Source Grounding}, \textit{Scenario-Condition Reasoning}, and \textit{Post-Edit Control and Validation}. In the second rollout, the agent restarts from the task under the same settings, and the Reminder mechanism examines each environment observation to decide whether to insert a Turn-local Reminder before the next action. Applying the same annotation protocol reveals fewer error events across all four categories (Fig.~\ref{fig:intro}). Post-hoc annotation audits completed trajectories, while runtime Reminder selection uses only the currently visible interaction prefix. This finding suggests that reliable long-horizon coding requires models to preserve task-level constraints across a trajectory and use local interaction feedback at appropriate decision points. A natural direction is therefore to provide process-level corrective information with explicit temporal scope.

However, coding-agent training relies on supervised imitation of successful trajectories or executable-outcome optimization~\citep{pan2025swegym,jain2025r2egym,wei2025swerl}. These signals improve task completion but do not indicate which constraints should persist or which observation should guide the next decision. At inference time, agents can adapt to feedback~\citep{yao2023react,chen2024selfdebug} or recover missed information through reflection and self-critique~\citep{shinn2023reflexion,madaan2023selfrefine,gou2024critic}. Runtime reminders require deployment-time construction and injection, consume context, and tie behavior to an agent scaffold. On-policy distillation trains students on self-generated outputs to reduce train--inference mismatch~\citep{agarwal2024onpolicy,gu2024minillm}. Related work uses context- and feedback-conditioned distillation to internalize behavior induced by auxiliary instructions or environment feedback~\citep{ye2026opcd,hubotter2026sdpo}. Applying these ideas to long-horizon coding leaves two challenges. First, training-time corrective information can encode persistent constraints and local corrections from observations relevant to one decision. Without explicit scopes, local corrections may be diluted in long contexts or propagated to unrelated decisions. Second, models without instruction-following adaptation may ignore constraints and feedback, so teacher reminders may not produce reliable distillation targets.

This raises a practical question: \textit{Can we internalize scope-specific corrective information during training so that a coding agent can operate without reminders at inference time?}

We address this with \textbf{On-Policy Reminder Distillation (OPRD)}, which combines Multi-Scope Reminder supervision with outcome-based routing on rollouts from a reminder-free student. For failed trajectories, we construct reminders from task, trajectory, and execution/validation records. A Global Reminder converts final-test diagnostics into trajectory-level guidance. A Turn-local Reminder is anchored to an observation and applies only to the next decision. This combines trajectory-level feedback with local correction without propagating local feedback to unrelated decisions. For reliable supervision, our two-stage cold start uses Coding SFT to establish coding capabilities and Instruction-Following (IF) SFT to improve responsiveness to constraints, feedback, and reminders. Both models are initialized from the IF checkpoint. Based on executable verification, OPRD routes successful rollouts to autoregressive self-imitation and failed rollouts to reminder-conditioned distillation without reinforcing erroneous decisions. For failures, the teacher re-evaluates student-generated prefixes under reminders, while the student matches the teacher distribution through Top-100 reverse KL. OPRD internalizes scope-specific corrections, enabling the student to act on corresponding constraints and feedback without reminders at inference time. We evaluate OPRD on \textsc{SWE-bench Verified}, \textsc{SWE-bench Pro}, and \textsc{SWE-bench Multilingual}~\citep{openai2024swebenchverified,deng2025swebenchpro,yang2025swesmith}. OPRD improves the benchmark-averaged task resolution rate by 11.3 points over the open-source Qwen3.5-4B base model. Under the same diagnostic protocol, the trained OPRD model reduces the aggregate error-event count across the four categories from 446 to 222, a 50.2\% reduction relative to the base model.

Our main contributions are threefold:
\begin{itemize}
    \item \textbf{Multi-Scope Reminder supervision for long-horizon coding.} We combine a Global Reminder that provides guidance across the trajectory with Turn-local Reminders that target only the next decision, enabling process-level supervision at complementary temporal scopes.
    \item \textbf{On-Policy Reminder Distillation with Outcome-Based Routing.} We propose OPRD, which learns from verified successful rollouts through autoregressive self-imitation and from failed rollouts through reminder-conditioned reverse KL distillation.
    \item \textbf{A two-stage supervised cold start for reliable reminder distillation.} Coding SFT followed by IF SFT provides a shared teacher--student initialization that improves responsiveness to constraints, execution feedback, and reminders, thereby providing a stronger starting point for subsequent reminder-conditioned distillation.
\end{itemize}
\section{Related Work}
\label{sec:related-work}

\begin{figure*}[htb]
    \centering
    \includegraphics[width=1.0\textwidth]{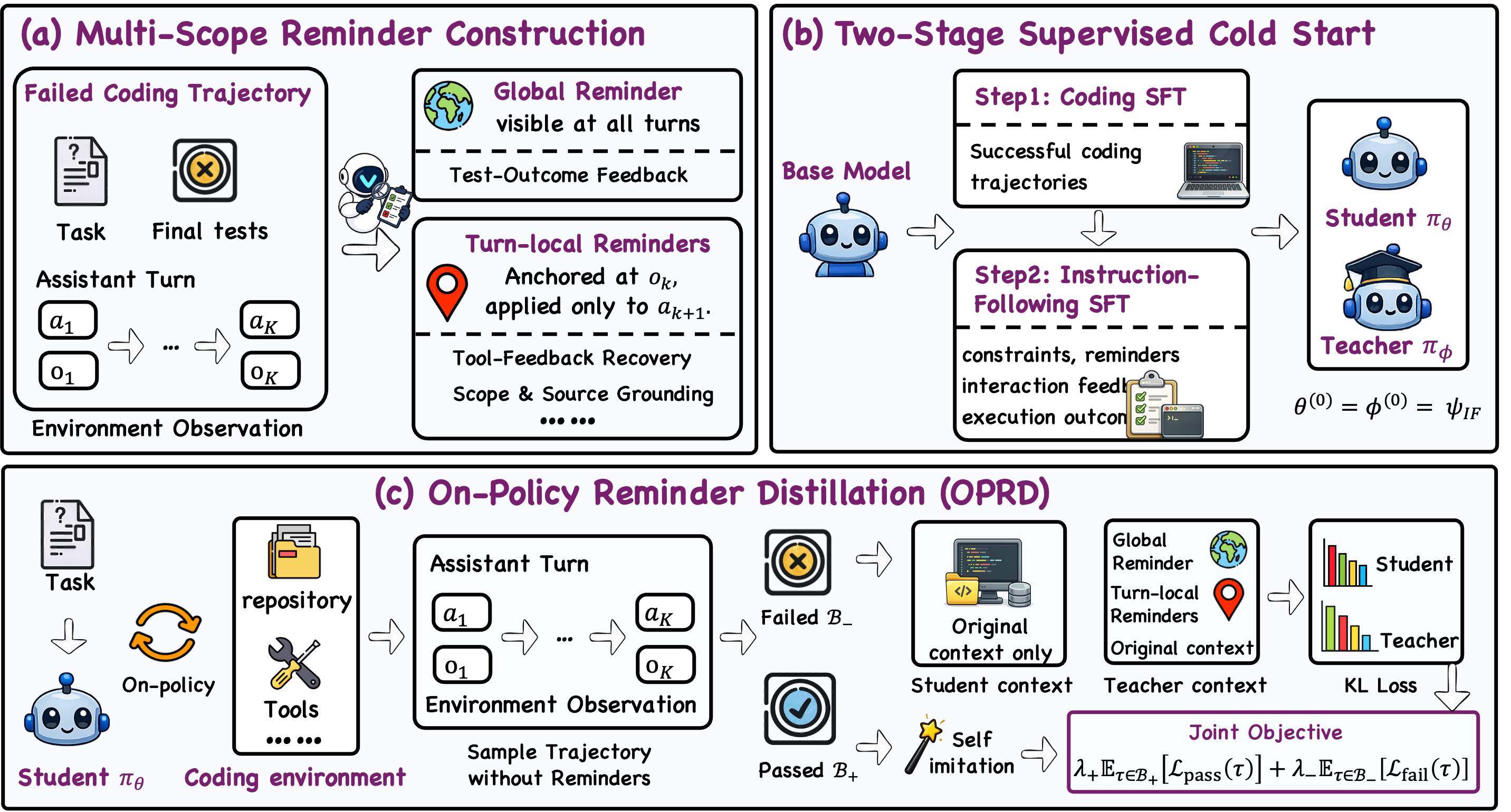}
    \caption{\textbf{Overview of OPRD.} OPRD constructs Global Reminders and observation-anchored Turn-local Reminders, and initializes the teacher and student through Coding SFT followed by IF SFT. During on-policy training, successful rollouts are optimized through autoregressive self-imitation, while failed rollouts receive reverse-KL supervision from a Reminder-conditioned teacher.}    
    \label{fig:oprd}
    \end{figure*}
\paragraph{Repository-Level Software Engineering Agents.}
Repository-level evaluation covers code completion~\citep{deng2024r2c2coder,yang2024execrepobench}, executable issue resolution~\citep{jimenez2024swebench}, broad agent capabilities~\citep{xu2025swecompass}, and multimodal web coding~\citep{lei2026webcompass}. \textsc{SWE-agent}, \textsc{AutoCodeRover}, and \textsc{Agentless} provide tool-based issue-resolution pipelines~\citep{yang2024sweagent,zhang2024autocoderover,xia2025agentless}. Training uses trajectories, verifiers, and generated or freshly collected tasks~\citep{pan2025swegym,jain2025r2egym,yang2025swesmith,badertdinov2025swerebench}, alongside executable-outcome optimization~\citep{wei2025swerl} and multi-stage coding-agent training~\citep{zhan2025katcoder}. Evaluation also considers engineering-constraint compliance beyond functional correctness~\citep{zhu2026anchor}.

\paragraph{Feedback-Guided Reasoning and Reflection.}
\textsc{ReAct}, \textsc{Reflexion}, and self-refinement use observations or feedback to guide actions and revisions~\citep{yao2023react,shinn2023reflexion,madaan2023selfrefine}. FTTT studies test-time feedback optimization~\citep{li2025fttt}; tool-grounded critique, search, and execution verification guide correction~\citep{gou2024critic,zhou2024lats,chen2024selfdebug,zhong2024ldb}. Complementary work explores code-critique evaluation~\citep{zhang2025codecriticbench}, failure diagnosis in agent traces~\citep{li2026codetracer}, visual feedback for web-code refinement~\citep{li2025relook}, and generative process rewards~\citep{xu2026flare}. Structured failure feedback also guides refinement in autonomous driving~\citep{luo2026elfvla}.

\paragraph{On-Policy and Privileged-Context Distillation.}
On-policy distillation supervises student-generated sequences~\citep{agarwal2024onpolicy}, with reverse/skew KL objectives~\citep{gu2024minillm,ko2024distillm} and entropy-aware or black-box extensions~\citep{jin2026entropyaware,ye2025blackbox}. Context and privileged-information distillation internalize behavior induced by auxiliary instructions or training-only information~\citep{snell2022distilling,zhao2026selfdistilled,penaloza2026privileged}. OPCD uses a context-conditioned teacher on student trajectories~\citep{ye2026opcd}. Reflective driving guidance can also be internalized through outcome-routed on-policy self-distillation~\citep{luo2026reflectvla}. OPRD studies this training--inference separation with explicitly scoped Reminders for long-horizon coding.

\section{Method}

Long-horizon coding agents must not only generate correct code, but also consistently follow user constraints throughout multi-turn interactions and adapt their behavior to tool responses and execution feedback. However, final verification outcomes provide limited process-level supervision for constraint adherence and feedback use at different scopes. To address this limitation, we propose \textbf{O}n-\textbf{P}olicy \textbf{R}eminder \textbf{D}istillation \textbf{(OPRD)}, which distills the behavior of a teacher conditioned on Multi-Scope Reminders into a student that operates without reminders at inference time. As illustrated in Fig.~\ref{fig:oprd}, we formulate the long-horizon coding task (Sec.~\ref{sec:problem}), then introduce Multi-Scope Reminders (Sec.~\ref{sec:reminder}) and a two-stage supervised cold start (Sec.~\ref{sec:if-init}), and present Multi-Scope Reminder-Conditioned Distillation (Sec.~\ref{sec:oprd}).

\subsection{Problem Formulation}
\label{sec:problem}

Given a repository-level coding task \(x\), an agent interacts with a code execution environment over multiple turns. We denote the \(k\)-th assistant turn by \(a_k\) and the resulting environment observation by \(o_k\). The complete interaction trajectory is
\begin{equation}
\tau=(x,a_1,o_1,\ldots,a_K,o_K).
\end{equation}
An executable verifier evaluates the final repository state and returns \(c(\tau)\in\{0,1\}\), where \(c(\tau)=1\) indicates that the trajectory passes verification and \(c(\tau)=0\) indicates failure. We denote the trainable student by \(\pi_\theta\) and the teacher by \(\pi_\phi\). The token sequence of the \(k\)-th assistant turn is \(a_k=(y_{k,1},\ldots,y_{k,L_k})\), where \(L_k\) is the number of tokens in that turn. The interaction history preceding the \(k\)-th assistant turn and the student context for generating its \(j\)-th token are defined as
\begin{equation}
H_k=(x,a_1,o_1,\ldots,a_{k-1},o_{k-1}),
\qquad
h_{k,j}^{S}=(H_k,y_{k,<j}),
\end{equation}
where \(y_{k,<j}\) denotes the token prefix preceding \(y_{k,j}\) within the current turn. System messages, user messages, and tool responses provide generation context but are not prediction targets. Training is restricted to valid assistant tokens, whose index set is
\begin{equation}
\Omega(\tau)
\subseteq
\{(k,j)\mid 1\leq k\leq K,\ 1\leq j\leq L_k\}.
\end{equation}

\subsection{Multi-Scope Reminders}
\label{sec:reminder}

The constraints and feedback received by a long-horizon coding agent operate over different temporal scopes. Task-level constraints, such as permitted modification scope, interface requirements, and output format, must hold throughout the trajectory. In contrast, test errors, tool responses, and file inspection results typically pertain to a specific decision point. Nevertheless, the final verifier compresses the entire trajectory into a binary outcome and cannot represent these distinct scopes. Treating all corrective information as a single trajectory-wide prompt may cause local errors to be diluted by the long context, whereas propagating local feedback to all subsequent turns may introduce irrelevant or conflicting supervision as the interaction state evolves. We therefore organize corrective information from a failed trajectory into a Multi-Scope Reminder. A Global Reminder provides persistent task-level guidance across turns, while a Turn-local Reminder is aligned with a specific decision point and highlights environmental feedback that the model ignored or misinterpreted at that point. For a failed trajectory \(\tau\), its Multi-Scope Reminder is represented as
\begin{equation}
R(\tau)=\left(r^{G},\{(k,r_k^{L})\}_{k\in\mathcal I_\tau}\right),
\qquad
\mathcal I_\tau\subseteq\{1,\ldots,K-1\},
\end{equation}
where \(r^{G}\) denotes the Global Reminder, \(r_k^{L}\) denotes the Turn-local Reminder anchored to environment observation \(o_k\), and \(\mathcal I_\tau\) is the set of observation indices associated with local corrective signals. The Global Reminder remains visible throughout the teacher's evaluation of assistant turns. In contrast, \(r_k^{L}\) is visible only when the teacher evaluates the assistant turn \(a_{k+1}\) following \(o_k\), and does not propagate to later turns. Instruction-Following (IF) SFT first strengthens the model's ability to understand and respond to explicit constraints and feedback. During OPRD, Multi-Scope Reminders serve as training-time privileged information available only to the teacher, and the behavior they induce is distilled into a student that does not observe reminders.

\paragraph{Reminder Construction.}
When a trajectory contains a correctable failure, we construct reminders post hoc from the task, trajectory, interaction feedback, and final test results. We determine the scope of each corrective signal and then assign its functional category. The Global Reminder contains one category, \emph{Test-Outcome Feedback}, which converts final-test diagnostics into trajectory-level guidance. Turn-local Reminders target correctable decisions identified from each environment observation and assistant action. They comprise four categories: \emph{Tool-Feedback Recovery} for responding to explicit tool feedback; \emph{Scope and Source Grounding} for evidence-grounded localization and modification; \emph{Scenario-Condition Reasoning} for task-specific execution conditions; and \emph{Post-Edit Control and Validation} for controlled post-edit verification. We retain grounded and actionable candidates, discarding ambiguous, redundant, unverifiable, reference-patch-revealing, or unsupported content. Detailed category definitions, generation prompts, trigger rules, and filtering procedures are provided in Appendix~\ref{app:reminder-construction}.

\subsection{Two-Stage Supervised Cold Start}
\label{sec:if-init}

The base model first acquires capabilities for repository understanding, tool interaction, and code editing through Coding SFT. However, coding competence alone does not ensure that the model will follow constraints and feedback that arise during an interaction. A standard coding model may still ignore user requirements, tool responses, test errors, or harness rules that are already present in the context. Given a base model \(\psi_0\), we first train on successful coding trajectories \(\mathcal D_{\mathrm{code}}\) to establish core agentic coding capabilities. We then use IF data \(\mathcal D_{\mathrm{IF}}\) to improve the model's responsiveness to user constraints, interaction feedback, execution outcomes, and reminders, yielding an IF checkpoint \(\psi_{\mathrm{IF}}\) that combines coding and instruction-following capabilities:
\begin{equation}
\psi_{\mathrm{code}}=\operatorname{SFT}(\psi_0,\mathcal D_{\mathrm{code}}),
\qquad
\psi_{\mathrm{IF}}=\operatorname{SFT}(\psi_{\mathrm{code}},\mathcal D_{\mathrm{IF}}).
\end{equation}

\subsection{Multi-Scope Reminder-Conditioned Distillation}
\label{sec:oprd}

OPRD uses shared initialization. 
The teacher \(\pi_\phi\) and student \(\pi_\theta\) use the same model architecture and are initialized with the parameters of the same IF checkpoint \(\psi_{\mathrm{IF}}\), such that \(\theta^{(0)}=\phi^{(0)}=\psi_{\mathrm{IF}}\).
During training, the teacher is periodically updated toward the student, while the student is optimized at every step. The teacher additionally receives reminders as training-time privileged information. For a task \(x\sim\mathcal D_{\mathrm{task}}\), the current student interacts with the coding environment without access to reminders to sample \(\tau\sim\pi_\theta(\cdot\mid x)\). As the student is updated, subsequent rollouts expose different states and errors, keeping reminder supervision aligned with the distribution visited by the current policy.

\paragraph{Outcome-Based Routing.}
After each rollout, an executable verifier determines whether it succeeds but does not provide fine-grained credit assignment. A successful rollout provides executable evidence that the student's current behavior can solve the task and can therefore serve directly as an autoregressive self-imitation signal. Applying the same objective to a failed rollout would instead increase the likelihood of erroneous decisions. OPRD thus routes successful rollouts to a self-imitation branch and failed rollouts to a reminder-conditioned distillation branch. For the current rollout batch \(\mathcal B\), we define \(\mathcal B_{+}=\{\tau\in\mathcal B\mid c(\tau)=1\}\) and \(\mathcal B_{-}=\{\tau\in\mathcal B\mid c(\tau)=0\}\).

\paragraph{Autoregressive Self-Imitation from Successful Rollouts.}
For each \(\tau\in\mathcal B_{+}\), we compute the autoregressive loss over all valid assistant tokens to increase the likelihood that the student reproduces successful behavior discovered through its own on-policy sampling:
\begin{equation}
\mathcal L_{\mathrm{pass}}(\tau)
=-\frac{1}{|\Omega(\tau)|}
\sum_{(k,j)\in\Omega(\tau)}
\log\pi_\theta(y_{k,j}\mid h_{k,j}^{S}).
\end{equation}

\paragraph{Reminder-Conditioned Distillation from Failed Rollouts.}
For each \(\tau\in\mathcal B_{-}\), reminders are constructed after the rollout and provided only to the teacher. The student always uses the original context without reminders. The teacher does not resample the trajectory. Instead, it evaluates the same student-generated token prefix \(y_{k,<j}\), ensuring that the distillation targets remain aligned with the states visited by the current policy. For the \(k\)-th assistant turn, the teacher history is defined as
\begin{equation}
\widetilde H_k^{T}
=
\begin{cases}
\operatorname{Inject}(H_k,r^{G},r_{k-1}^{L}),
& \substack{k\geq 2,\\ k-1\in\mathcal I_\tau},\\
\operatorname{Inject}(H_k,r^{G}),
& \text{otherwise}.
\end{cases}
\end{equation}
The corresponding teacher token context is \(h_{k,j}^{T}=(\widetilde H_k^{T},y_{k,<j})\). The teacher augments the student's original context with the reminders active at the current scope. The Global Reminder is visible at every assistant turn, whereas \(r_{k-1}^{L}\) is injected only when the teacher evaluates \(a_k\) and is removed from subsequent turns. Reminder scope is determined by its visibility in the teacher context, while the loss for a failed trajectory still covers all valid assistant tokens in \(\Omega(\tau)\). Within this unified distillation signal, the Global Reminder provides task-level correction across turns, while the Turn-local Reminder provides fine-grained guidance at the corresponding decision point. Let \(\mathcal V\) denote the full vocabulary, and let \(z_\theta\) and \(z_\phi\) denote the student and teacher logits, respectively. For \(v\in\mathcal V\), their token distributions at temperature \(\mathcal T\) are
\begin{equation}
\begin{split}
p_{\theta,k,j}^{(\mathcal T)}(v)
&=\operatorname{softmax}\!\left(
\frac{z_\theta(h_{k,j}^{S})}{\mathcal T}
\right)_v,\\
p_{\phi,k,j}^{(\mathcal T)}(v)
&=\operatorname{softmax}\!\left(
\frac{z_\phi(h_{k,j}^{T})}{\mathcal T}
\right)_v.
\end{split}
\end{equation}

We then compute the reverse Kullback--Leibler (KL) divergence over the Top-K-vocabulary distributions:
\begin{equation}
\mathcal L_{\mathrm{fail}}(\tau)
=\frac{\mathcal T^2}{|\Omega(\tau)|}
\sum_{(k,j)\in\Omega(\tau)}
D_{\mathrm{KL}}
\left[
p_{\theta,k,j}^{(\mathcal T)}
\;\middle\|\;
\operatorname{sg}\!\left[p_{\phi,k,j}^{(\mathcal T)}\right]
\right],
\end{equation}
where \(\operatorname{sg}\) stops gradients through the teacher branch and \(\mathcal T^2\) compensates for gradient-scale changes from temperature scaling. Reverse KL weights discrepancies by the student distribution, prioritizing corrections to tokens to which the student assigns high probability in visited states.

\paragraph{Joint Objective and Inference.}
After training, only the student is deployed. The teacher, the reminder-construction module, and all reminders are removed at inference time. Let \(\lambda_{+}\) and \(\lambda_{-}\) control the relative strengths of self-imitation on successful rollouts and reminder distillation on failed rollouts, respectively. The OPRD objective is
\begin{equation}
\mathcal L_{\mathrm{OPRD}}
=\lambda_{+}\mathbb{E}_{\tau\in\mathcal B_{+}}[\mathcal L_{\mathrm{pass}}(\tau)]
+\lambda_{-}\mathbb{E}_{\tau\in\mathcal B_{-}}[\mathcal L_{\mathrm{fail}}(\tau)].
\end{equation}

\section{Experiments}
\label{sec:experiments}

\subsection{Experimental Setup}
\label{sec:experimental-setup}

\paragraph{Models and Training Data.}
\label{sec:training-data}
We use Qwen3.5-4B and Qwen3.5-9B as base models~\citep{qwen2026qwen35}. Training consists of three stages. Coding SFT uses 8,000 verified successful trajectories, including 7,262 from \textsc{SWE-rebench V2}~\citep{badertdinov2026swerebenchv2} and 738 from Multi-SWE-RL~\citep{zan2025multiswebench}. IF SFT then uses 2,000 reminder-following examples from \textsc{SWE-rebench V2}. OPRD operates on a pool of 3,630 \textsc{SWE-rebench V2} tasks and 370 Multi-SWE-RL tasks, from which the student generates reminder-free rollouts online. In the OPRD variant without IF SFT, both teacher and student are initialized from the Coding SFT checkpoint while all other settings remain unchanged.

Table~\ref{tab:training-data-summary} summarizes the three stage-specific data configurations. Coding SFT and IF SFT operate on trajectories or reminder-following examples, whereas OPRD uses a task pool from which the current student generates new rollouts online. The OPRD counts therefore describe available training tasks rather than a fixed set of trajectories.

\begin{table}[t]
    \centering
    \small
    \begin{tabular}{lcrr}
        \toprule
        Stage & Unit &
        \shortstack{\textsc{SWE-rebench}\\V2} &
        \shortstack{Multi-\\SWE-RL} \\
        \midrule
        Coding SFT & Trajectory & 7,262 & 738 \\
        IF SFT & Example & 2,000 & 0 \\
        OPRD & Task & 3,630 & 370 \\
        \bottomrule
    \end{tabular}
    \caption{Training data used at each stage.}
    \label{tab:training-data-summary}
\end{table}

\begin{figure}[t]
    \centering
    \includegraphics[width=\columnwidth]{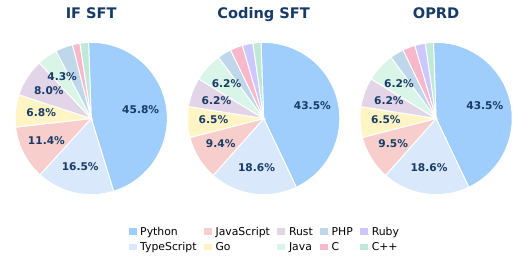}
    \caption{Programming-language distributions for IF SFT, Coding SFT, and the OPRD task pool. Percentages are computed from 2,000 examples, 8,000 examples, and 4,000 examples, respectively. The same language-color mapping is used in all three pies.}
    \label{fig:training-language-distribution}
\end{figure}

Figure~\ref{fig:training-language-distribution} reports the programming-language composition of the three stages. Python is the largest source in every stage (43.5--45.8\%), followed by TypeScript and JavaScript. Coding SFT and the OPRD task pool use nearly identical proportions by construction, while the IF set is slightly more concentrated in Python, JavaScript, and Rust and contains no Ruby example. The distribution nevertheless covers nine languages in IF SFT and all ten languages in Coding SFT and OPRD.

\paragraph{Implementation Details.}
Training implementation and hyperparameters are detailed in Appendix~\ref{app:training-hyperparameters}.

\paragraph{Evaluation Setup.}
\label{sec:evaluation-protocol}
We evaluate task resolution rate on \textsc{SWE-bench Verified}, \textsc{SWE-bench Pro}, and \textsc{SWE-bench Multilingual}~\citep{openai2024swebenchverified,deng2025swebenchpro,yang2025swesmith}. All models use the same agent harness and inference configuration. Except for the controlled Reminder interventions in the scope and functional-category analyses, all main results and training ablations use reminder-free inference.

All models are evaluated with a shared Claude Code agent scaffold and the executable verifier associated with each benchmark. Every subtask is allocated 16 CPU cores and 32,768 MiB of memory, with a maximum of 200 agent turns. Decoding uses temperature 1.0 and top-\(p\) 0.95. The vLLM deployment permits at most 128 concurrent sequences, a maximum model length of 262,144 tokens, and 8,192 batched tokens.

\subsection{Main Results}
\label{sec:main-results}

\begin{table*}[t]
    \centering
    \small
    \begin{tabular}{lcccc}
        \toprule
        \multicolumn{1}{c}{%
            \multirow{2}{*}{Training Configuration}
        } &
        \textsc{SWE-bench} &
        \textsc{SWE-bench} &
        \textsc{SWE-bench} &
        \multirow{2}{*}{Average (\%)} \\
        &
        Verified (\%) &
        Pro (\%) &
        Multilingual (\%) &
        \\
        \midrule
        Qwen3.5-4B
        & 50.4 & 24.2 & 35.3 & 36.6 \\

        \(+\) Coding SFT
        & 53.6 & 25.6 & 38.7 & 39.3 \\

        \(+\) Coding SFT \(\rightarrow\) IF SFT
        & 55.8 & 26.7 & 42.3 & 41.6 \\

        \(+\) Coding SFT \(\rightarrow\) OPRD (w/o IF SFT)
        & \underline{58.4}
        & \underline{28.3}
        & \underline{46.3}
        & \underline{44.3} \\

        \rowcolor[gray]{0.92}
        \(+\) Coding SFT \(\rightarrow\) IF SFT
        \(\rightarrow\) OPRD (Ours)
        & \textbf{61.2}
        & \textbf{30.5}
        & \textbf{52.0}
        & \textbf{47.9} \\
        \midrule
        Qwen3.5-9B
        & 59.6 & 32.3 & 41.3 & 44.4 \\

        \(+\) Coding SFT
        & 61.4 & 34.7 & 47.7 & 47.9 \\

        \(+\) Coding SFT \(\rightarrow\) IF SFT
        & 64.0 & 36.7 & 50.7 & 50.5 \\

        \(+\) Coding SFT \(\rightarrow\) OPRD (w/o IF SFT)
        & \underline{67.2} & \underline{38.9} & \underline{53.7} & \underline{53.3} \\

        \rowcolor[gray]{0.92}
        \(+\) Coding SFT \(\rightarrow\) IF SFT \(\rightarrow\) OPRD (Ours)
        & \textbf{70.0} & \textbf{40.9} & \textbf{56.7} & \textbf{55.9} \\
        \bottomrule
    \end{tabular}
    \caption{Main results with Qwen3.5-4B and Qwen3.5-9B on \textsc{SWE-bench Verified}, \textsc{SWE-bench Pro}, and \textsc{SWE-bench Multilingual}. We report task resolution rates (\%) and their unweighted average. Within each backbone, the best and second-best results in each column are shown in bold and underlined, respectively.}
    \label{tab:main-results}
    \label{tab:qwen9b-results}
\end{table*}


\paragraph{OPRD provides consistent gains across software engineering benchmarks.}
Table~\ref{tab:main-results} compares the base and supervised checkpoints with the OPRD variants to assess whether Reminder-conditioned process supervision transfers to reminder-free inference. The complete training recipe reaches an average resolution rate of 47.9, improving over the Qwen3.5-4B base model by 11.3 points and over the IF SFT checkpoint by 6.3 points. Gains are consistent across all three benchmarks, with the largest improvement on \textsc{SWE-bench Multilingual}. Removing IF SFT before OPRD reduces the average by 3.6 points. This gap suggests that coding adaptation alone is insufficient for fully exploiting Reminder-conditioned supervision, whereas IF SFT provides a more suitable shared initialization for teacher and student. Because all entries in Table~\ref{tab:main-results} use reminder-free inference, the improvement does not depend on test-time Reminder injection.

\paragraph{OPRD gains persist with the 9B backbone.}
\label{sec:qwen9b-results}
As shown in Table~\ref{tab:qwen9b-results}, the complete Qwen3.5-9B training recipe reaches an average resolution rate of 55.9, improving over the open-source checkpoint by 11.5 points and over the IF SFT checkpoint by 5.4 points. The improvement over the open-source model is consistent on \textsc{SWE-bench Verified}, \textsc{SWE-bench Pro}, and \textsc{SWE-bench Multilingual}, with gains of 10.4, 8.6, and 15.4 points, respectively. Removing IF SFT before OPRD lowers the average by 2.6 points, showing that the supervised instruction-following preparation remains useful at the larger model scale.

\begin{figure}[!t]
    \centering
    \includegraphics[width=\columnwidth]{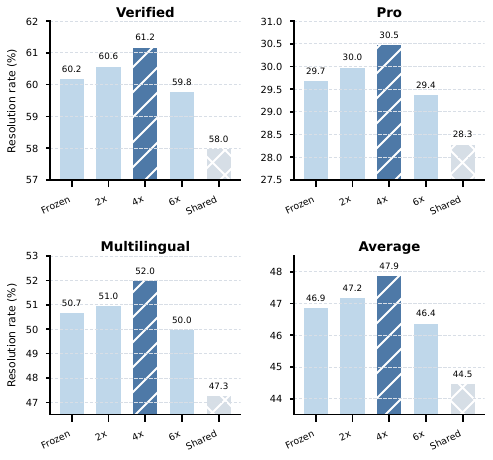}
    \caption{Effect of the teacher refresh schedule on Qwen3.5-4B. We compare Frozen Teacher, Periodic Refresh (\(2\times\), \(4\times\), and \(6\times\)), and Shared Parameters on three benchmarks and their unweighted average.}
    \label{fig:teacher-refresh}
\end{figure}

\begin{figure*}[!t]
    \centering
    \includegraphics[width=\textwidth]{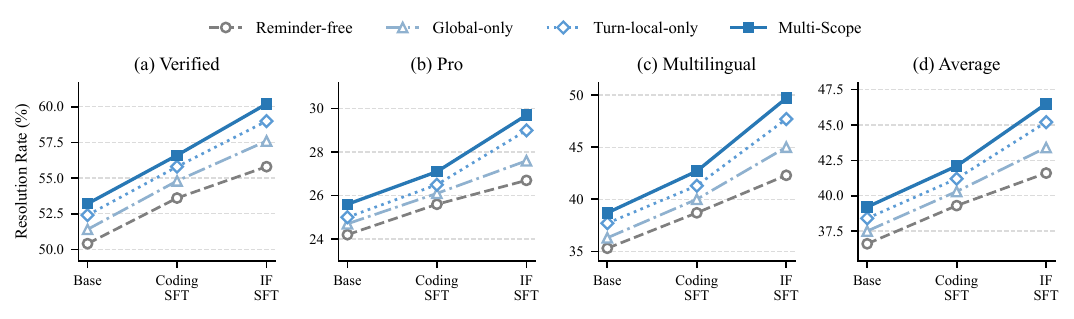}
    \caption{Effect of Reminder scope and supervised preparation on Qwen3.5-4B. We compare reminder-free, Global-only, Turn-local-only, and Multi-Scope Reminder conditioning for the Base, Coding SFT, and IF SFT checkpoints.}
    \label{fig:reminder-responsiveness}
\end{figure*}

\begin{figure}[!t]
    \centering
    \includegraphics[width=0.95\columnwidth]{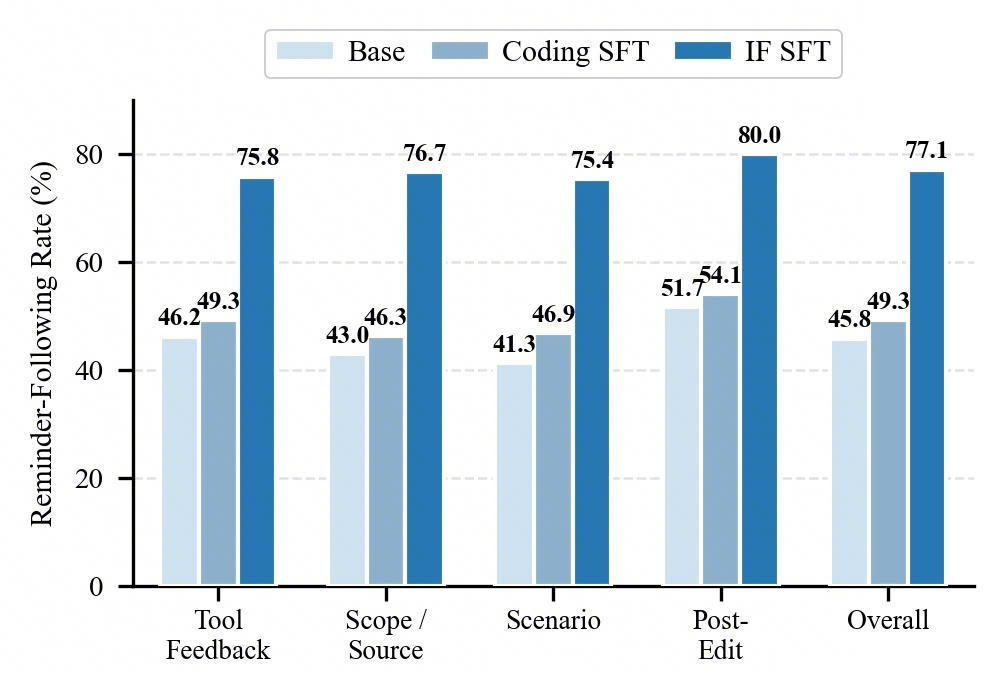}
    \caption{Reminder-following rates on 200 \textsc{SWE-rebench} V2 validation tasks. After a category-specific error is detected, the corresponding Reminder is inserted at the relevant interaction point, and the subsequent behavior is judged by whether it corrects or avoids that error. Each task--category pair contributes at most one successful follow event. \emph{Overall} aggregates the four functional categories.}
    \label{fig:if-reminder-following}
\end{figure}

\begin{figure}[!t]
    \centering
    \includegraphics[width=\columnwidth]{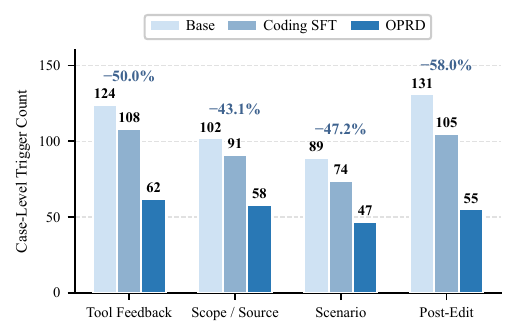}
    \caption{Diagnostic error reduction on 200 \textsc{SWE-rebench} V2 validation tasks. Bars show case-level trigger counts for Qwen3.5-4B (Base), Coding SFT, and OPRD. Percentages report relative reductions from Qwen3.5-4B to OPRD.}
    \label{fig:diagnostic-error-reduction}
\end{figure}

\begin{figure*}[!t]
    \centering
    \includegraphics[width=\textwidth]{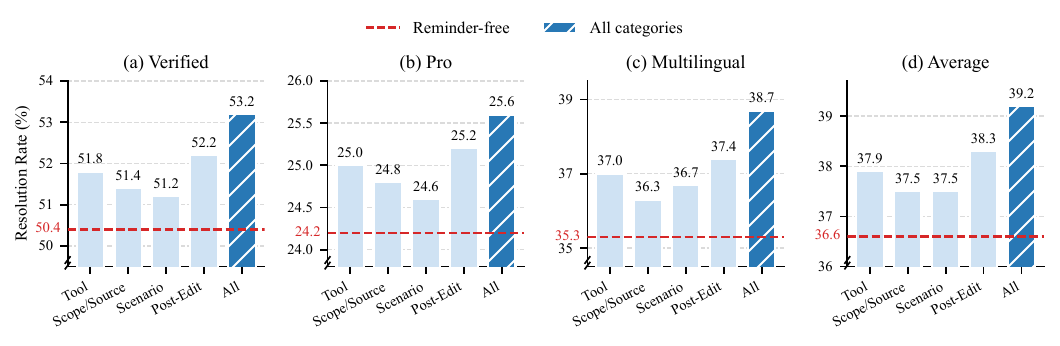}
    \caption{Effect of functional Reminder categories on the Qwen3.5-4B base checkpoint. We evaluate Tool-Feedback Recovery, Scope and Source Grounding, Scenario-Condition Reasoning, and Post-Edit Control and Validation, together with their combination. The dashed line denotes reminder-free performance.}
    \label{fig:reminder-categories}
\end{figure*}

\subsection{Ablation Study}
\label{sec:ablation-study}
\label{sec:more-analysis}

We examine the distillation design, loss balance, teacher refresh, supervised preparation, and Reminder scope and functional categories, followed by a process-level error analysis. Training ablations use the same task pool and on-policy rollout protocol; Reminder interventions are compared under shared inference settings.

\label{sec:variant-definitions}
The OPRD variant without IF SFT initializes both the teacher and student from the Coding SFT checkpoint and otherwise retains the complete OPRD configuration. In the distillation-formulation ablation, All-Rollout OPSD applies the selected KL objective to both successful and failed on-policy rollouts. Outcome-Routed OPRD instead applies autoregressive self-imitation to verified successes and Multi-Scope Reminder-conditioned KL to failures. The compared rows use the same training tasks, on-policy rollout budget, Top-100 support, and distillation temperature.

The teacher-refresh ablation compares a frozen teacher, periodic refresh schedules, and shared teacher--student parameters. The frozen teacher remains at its initial checkpoint. Our four-refresh setting copies the current student parameters to the teacher at 20\%, 40\%, 60\%, and 80\% of training. Under shared parameters, the teacher context is evaluated with the current student parameters at every update while gradients remain stopped through the teacher branch.

\begin{table*}[t]
    \centering
    \small
    \begin{tabular}{lcccc}
        \toprule
        \multicolumn{1}{c}{\multirow{2}{*}{Setting}} &
        \textsc{SWE-bench} &
        \textsc{SWE-bench} &
        \textsc{SWE-bench} &
        \multirow{2}{*}{Average (\%)} \\
        &
        Verified (\%) &
        Pro (\%) &
        Multilingual (\%) &
        \\
        \midrule

        \multicolumn{5}{l}{%
            \textit{KL Support and Distillation Temperature}
        } \\
        Full Vocabulary, \(\mathcal T=0.6\)
        & 59.8 & 29.4 & 50.3 & 46.5 \\

        Top-100, \(\mathcal T=0.6\)
        & 60.2 & 29.7 & 50.7 & 46.9 \\

        Top-100, \(\mathcal T=1.0\) (Ours)
        & \textbf{61.2}
        & \textbf{30.5}
        & \textbf{52.0}
        & \textbf{47.9} \\

        Top-100, \(\mathcal T=1.5\)
        & \underline{60.4}
        & \underline{29.8}
        & \underline{51.0}
        & \underline{47.1} \\

        \midrule

        \multicolumn{5}{l}{%
            \textit{Outcome-Based Routing and KL Direction}
        } \\
        All-Rollout OPSD, Forward KL
        & 58.8 & 28.6 & 48.3 & 45.2 \\

        All-Rollout OPSD, Reverse KL
        & 59.6 & 29.1 & 49.3 & 46.0 \\

        Outcome-Routed OPRD, Forward KL
        & \underline{60.2}
        & \underline{29.7}
        & \underline{50.3}
        & \underline{46.7} \\

        Outcome-Routed OPRD, Reverse KL (Ours)
        & \textbf{61.2}
        & \textbf{30.5}
        & \textbf{52.0}
        & \textbf{47.9} \\
        \bottomrule
    \end{tabular}
    \caption{Ablations of KL support, distillation temperature,
    outcome-based routing, and KL direction with Qwen3.5-4B.
    Scores are task resolution rates (\%), and \textit{Average} is the
    unweighted mean across the three benchmarks. The best and second-best
    results within each group are shown in bold and underlined, respectively.}
    \label{tab:distillation-ablation}
\end{table*}

\paragraph{Effect of KL support and distillation temperature.}
The KL-support ablation compares full-vocabulary and Top-100 supports and varies the distillation temperature to identify an effective distributional target for OPRD. At the shared temperature of \(\mathcal T=0.6\), Top-100 and full-vocabulary KL perform similarly, with Top-100 obtaining a slightly higher average score. This indicates that a sparse high-probability support is sufficient for effective distillation in this setting. Among the Top-100 variants, \(\mathcal T=1.0\) performs best with an average score of 47.9, while both lower and higher temperatures perform worse, suggesting that a moderate temperature balances teacher preferences with useful alternatives.

\paragraph{Effect of outcome-based routing and KL direction.}
A paired comparison of All-Rollout OPSD and Outcome-Routed OPRD under forward and reverse KL isolates the effects of outcome-based routing and KL direction. Reverse KL outperforms forward KL under both training strategies. At a fixed KL direction, OPRD consistently improves over OPSD, with a 1.9-point gain under reverse KL. These results support assigning verified successes to autoregressive self-imitation and failures to Reminder-conditioned distillation, rather than applying a single KL objective uniformly to all rollouts. Reverse KL further focuses the update on candidates that the student assigns high probability but the Reminder-conditioned teacher considers inappropriate.



\begin{table}[t]
    \centering
    \small
    \begin{tabular}{ccccc}
        \toprule
        \(\lambda_{+}:\lambda_{-}\) &
        Ver. & Pro & Multi. & Avg. \\
        \midrule
        0.25:1 & 59.6 & 28.7 & 49.7 & 46.0 \\
        0.5:1  & 60.4 & 29.5 & 51.0 & 47.0 \\
        \textbf{1:1~(Ours)} & \textbf{61.2} & \textbf{30.5} & \textbf{52.0} & \textbf{47.9} \\
        2:1    & 60.8 & 29.8 & 51.3 & 47.3 \\
        4:1    & 59.8 & 28.9 & 50.3 & 46.3 \\
        \bottomrule
    \end{tabular}
    \caption{Ablation of the relative weights assigned to successful-trajectory autoregressive self-imitation (\(\lambda_{+}\)) and failed-trajectory Reminder-conditioned distillation (\(\lambda_{-}\)). Columns report \textsc{SWE-bench Verified}, Pro, Multilingual, and their average.}
    \label{tab:lambda-balance}
\end{table}

\paragraph{Balancing successful and failed trajectories.}
The balanced \(1{:}1\) setting performs best on all three benchmarks and obtains a 47.9 average. Reducing \(\lambda_{+}\) over-concentrates training on correcting failed states and weakens retention of successful behavior already discovered by the current policy. Increasing \(\lambda_{+}\) gradually shifts the objective toward successful-trajectory self-imitation and underuses the process supervision supplied by Reminders on failed states. \textsc{SWE-bench Pro} and \textsc{SWE-bench Multilingual} are more sensitive to this imbalance because they contain more tasks that require correction from tool feedback and cross-file context.

\paragraph{Effect of the teacher refresh schedule.}
Varying the teacher refresh schedule examines the trade-off between adapting to the evolving student and maintaining a stable distillation target. As shown in Fig.~\ref{fig:teacher-refresh}, periodic refreshes exhibit a non-monotonic trend, with four evenly spaced refreshes performing best on every benchmark and reaching an average resolution rate of 47.9. This pattern suggests that a frozen teacher may become misaligned with states visited by the evolving student, whereas overly frequent refreshes may reduce target stability. The Shared Parameters setting performs worst, further supporting the importance of a relatively stable teacher target for Reminder-conditioned distillation.

\paragraph{Effect of Reminder scope.}
A controlled comparison of four Reminder conditions across the Base, Coding SFT, and IF SFT checkpoints measures Reminder utility and model responsiveness. Both Global-only and Turn-local-only conditioning improve every checkpoint on all benchmarks, while combining them consistently performs best. Turn-local Reminders generally provide larger gains, while the additional improvement from Global Reminders shows that persistent cross-turn guidance complements feedback aligned with a specific environment observation. The average benefit of Multi-Scope Reminders grows across preparation stages and reaches 4.9 points after IF SFT, indicating that IF SFT improves the model's ability to use corrective information at different scopes.

\paragraph{Effect of IF SFT on Reminder responsiveness.}
\label{sec:if-reminder-following}
As shown in Fig.~\ref{fig:if-reminder-following}, the aggregate Reminder-following rate increases from 45.8 for the open-source base model to 49.3 after Coding SFT, and then to 77.1 after IF SFT. Relative to Coding SFT, IF SFT improves the rate by 27.8 points overall and by 25.9--30.4 points across the four functional categories. Its category-level rates are consistently high, ranging from 75.4 for Scenario-Condition Reasoning to 80.0 for Post-Edit Control and Validation. The largest gain over Coding SFT occurs for Scope and Source Grounding, which increases from 46.3 to 76.7. These results isolate behavioral adherence to an inserted Reminder rather than end-to-end task resolution, and support using the IF checkpoint as the shared teacher--student initialization for OPRD.

\paragraph{Effect of functional Reminder categories.}
Grouping Reminders by content rather than temporal scope evaluates the contribution of each functional category. All four categories in Fig.~\ref{fig:reminder-categories} outperform reminder-free evaluation, with Post-Edit Control and Validation providing the strongest individual result. Enabling all categories raises the average resolution rate from 36.6 to 39.2, suggesting that they provide complementary corrective signals and broader coverage when combined.

\paragraph{Process-level error reduction.}
Applying the same four-category taxonomy as Fig.~\ref{fig:intro} tests whether OPRD's task-level gains coincide with fewer process errors. Each checkpoint generates one reminder-free trajectory for the same 200 \textsc{SWE-rebench} V2 validation tasks under identical inference settings, and we record case-level triggers from the resulting trajectories. As shown in Fig.~\ref{fig:diagnostic-error-reduction}, the total number of triggers falls from 446 for Qwen3.5-4B to 378 after Coding SFT and 222 after OPRD, a 50.2\% reduction relative to Qwen3.5-4B. Trigger counts decline by 43.1--58.0\% across categories, with the largest decrease in Post-Edit Control and Validation. These results show that OPRD improves both final task resolution and intermediate process behavior.

\FloatBarrier

\section{Conclusion}

Long-horizon coding agents must preserve task constraints and respond to interaction feedback while producing correct code. Our diagnostic analysis highlights process-level failures throughout repository-level problem solving. We introduce OPRD, which uses Multi-Scope Reminders as training-time privileged information and applies outcome-based routing to on-policy rollouts. A Global Reminder converts final-test diagnostics into trajectory-level guidance, while Turn-local Reminders anchor observation-derived corrections to the next assistant decision. OPRD routes successful rollouts to autoregressive self-imitation and failed rollouts to Reminder-conditioned reverse-KL distillation. Its two-stage supervised cold start, consisting of Coding SFT followed by Instruction-Following SFT, improves responsiveness to these corrective signals. Experiments show that OPRD improves task resolution and reduces adherence failures. By internalizing trajectory-level feedback and local corrections into model parameters, OPRD enables long-horizon coding agents to operate without Reminders at inference time.

\bibliography{aaai2027}


\clearpage
\appendix
\section{Appendix} \label{appendix:appendix} 
In this appendix, we provide training implementation and hyperparameters (Section~\ref{app:training-hyperparameters}), together with supplementary materials on Reminder construction and diagnostic annotation. \textbf{Section~\ref{app:reminder-construction}} details the construction procedures, functional taxonomy, Global Reminder evidence processing, and complete runtime category-judge prompt. \textbf{Section~\ref{app:diagnostic-error-annotation}} describes the deterministic and model-based annotation protocol used to analyze coding-agent errors. Training data, evaluation settings, results, and ablations are presented in Section~\ref{sec:experiments}.

\section{Training Implementation and Hyperparameters}
\label{app:training-hyperparameters}
\label{sec:training-hyperparameters}
We implement OPRD using VeOmni~\citep{ma2025veomni} with FlashAttention 2~\citep{dao2023flashattention2}, fused MoE operators, FSDP2~\citep{zhao2023pytorchfsdp}, and Ulysses sequence parallelism~\citep{jacobs2023deepspeedulysses}. Distributed checkpoints are stored with DCP. Table~\ref{tab:training-hyperparameters} lists the supplied Qwen3.5-4B configuration.

The main Qwen3.5-4B configuration uses a maximum sequence length of 131,072, a global batch size of 32, 3 training epochs, and a peak learning rate of \(1\times10^{-5}\) with cosine decay. OPRD uses Top-100 reverse KL at temperature \(\mathcal T=1.0\).

\begin{table*}[t]
    \centering
    \caption{Training configuration for the Qwen3.5-4B OPRD experiment.}
    \label{tab:training-hyperparameters}
    \small
    \begin{tabular}{llll}
        \toprule
        Parameter & Value & Parameter & Value \\
        \midrule
        Model & Qwen3.5-4B &
        Maximum sequence length & 131,072 \\

        Training framework & VeOmni &
        LR start & 0 \\

        Epochs & 3 &
        GPU & 16 NVIDIA H20 \\

        Global batch size & 32 &
        Micro batch size & 1 \\

        Peak learning rate & \(1\times10^{-5}\) &
        Minimum learning rate & \(1\times10^{-6}\) \\

        LR schedule & Cosine &
        Warmup ratio & 0.005 \\

        LR decay ratio & 1 &
        Weight decay & 0 \\

        Distillation temperature & 1.0 &
        KL direction & Reverse \\

        KL vocabulary support & Top-100 &
        Teacher refresh & 20/40/60/80\% \\

        Self-imitation weight \(\lambda_{+}\) & 1.0 &
        Distillation weight \(\lambda_{-}\) & 1.0 \\

        Expert parallel size & 1 &
        Ulysses size & 4 \\

        Data-parallel shard size & 2 &
        Attention implementation & FlashAttention 2 \\
        \bottomrule
    \end{tabular}
\end{table*}

\paragraph{Teacher refresh.}
The teacher is not optimized by gradient descent. At 20\%, 40\%, 60\%, and 80\% of OPRD training, its parameters are replaced by the current student parameters and then kept fixed until the next refresh. This schedule yields four teacher refreshes over a complete run.

\paragraph{Top-\(k\) KL support.}
The main configuration evaluates reverse KL on one shared Top-100 support at each valid assistant-token position. After temperature scaling, we compute a joint importance score \(s(v)=\max\{\log p_{\theta}(v),\log p_{\phi}(v)\}\) for every vocabulary item \(v\), where \(p_{\theta}\) and \(p_{\phi}\) denote the student and stop-gradient teacher distributions. We select the 100 tokens with largest \(s(v)\), then renormalize both distributions on exactly this common index set before evaluating reverse KL. Selecting by the pointwise maximum retains tokens considered plausible by either model while keeping the support size fixed.

The main loss uses \(\lambda_{+}=1\) for autoregressive self-imitation on verified successful trajectories and \(\lambda_{-}=1\) for Reminder-conditioned distillation on failed trajectories.

\section{Reminder Construction Details}
\label{app:reminder-construction}

\subsection{Construction Procedure and Functional Taxonomy}

Reminder construction and Reminder categorization describe two distinct parts of our pipeline. Construction specifies how a candidate is obtained, filtered, inserted, and verified, whereas the functional taxonomy specifies which coding-agent failure the retained Reminder addresses. These two attributes are independent of the temporal scope used by OPRD: a Global Reminder provides persistent final-test guidance, while the runtime construction procedure below is used to construct and evaluate observation-anchored Turn-local Reminders.

We construct runtime Reminder candidates from 
\textsc{SWE-rebench} V2 tasks and naturally occurring coding-agent trajectories. For each trajectory, we align the task requirement, assistant action, tool call, tool result, next assistant action, and subsequent validation outcome. A candidate is retained only when its trigger is observable without hidden correctness labels, its desired next action is local and behaviorally verifiable, and its text neither reveals a reference patch nor invents a new task requirement. Correctness is determined separately by the executable ReBench evaluator; a trajectory alone is used only as behavioral evidence.

\paragraph{Passive and active construction modes.}
A \emph{passive} Reminder is appended immediately after a tool result exposes a concrete failure, ambiguity, scope drift, or validation signal. It is one-shot: the next assistant decision observes it, but it does not automatically persist to later turns. An \emph{active} Reminder is evaluated at a controlled pre-action boundary. We first obtain a probe action, use it only to determine whether the trigger fires, discard it when an intervention is required, insert the Reminder into the same visible state, and reroll the action. The active verifier then evaluates the rerolled tool choice, arguments, target path, edit scope, and immediate follow-up behavior. Thus passive construction mines recovery from observed feedback, whereas active construction provides a matched intervention for measuring causal behavior change.

\begin{table*}[t]
    \centering

    \small
    \caption{The four reporting classes used for functional Reminder analysis. The mapping is a fixed partition of the 32 original rubric categories.}

    \begin{tabular}{cllc}
        \toprule
        ID & Functional Class & Original Rubric Source & Count \\
        \midrule
        R1 & Tool-Feedback Recovery & Deterministic hard rules & 10 \\
        R2 & Scope and Source Grounding & Scope/patch and source/path slots & 10 \\
        R3 & Scenario-Condition Reasoning & Scenario/condition slot & 5 \\
        R4 & Post-Edit Control and Validation & Post-edit drift and validation/wrap-up slots & 7 \\
        \midrule
        & Total & 10 hard rules and 22 non-hard strategies & 32 \\
        \bottomrule

    \end{tabular}
    \label{tab:reminder-taxonomy-summary}
\end{table*}

\paragraph{Runtime selection and insertion.}
At each legal insertion boundary, deterministic hard rules are matched against the latest tool result before a model judge considers non-hard contextual strategies. Non-hard candidates must reach confidence 0.90 and satisfy their phase gate. We then apply a per-family cap, a per-case budget, repeated-text removal, and hard-error-signature deduplication. Post-edit candidates require a visible edit, and validation-wrap-up candidates require concrete test, build, compiler, diff, formatting, or completion evidence. If all candidates fail these gates, no Reminder is inserted; the system does not use a fallback category. Hard-rule text is copied from the rubric, and every inserted Reminder is serialized after, rather than inside, the corresponding tool response.

We separately record whether trajectory capture completed, a Reminder was inserted, the required local behavior was followed, and the final source-only patch passed executable verification. These quantities use different denominators and are not substituted for one another.

\subsection{Global Reminder Construction and Evidence Judge}
\label{app:global-reminder}

Global Reminders are constructed only after a complete coding attempt has been rejected by executable verification. Unlike a Turn-local Reminder, which is anchored to one observation and governs the next decision, a Global Reminder summarizes the previous patch and its verifier artifacts and is placed at the input of the next full retry. A candidate is eligible only if it contains a non-empty patch, receives zero benchmark reward, is not marked non-retryable, is successfully applied in the verification environment, and is not accepted by the benchmark. These gates exclude empty or unapplied patches, successful attempts, and infrastructure-only records.

The evidence judge parses the benchmark result JSON and raw test log, records available pass/pending/fail counts and touched files, deduplicates failures with a shared root cause, and selects at most five signals. Root-cause-bearing interface, argument, syntax, reference, symbol, path, and signature failures are ranked before assertion, command, and generic failures. The resulting \texttt{verification\_feedback} states what failed and where the evidence came from, but does not reveal a reference patch or prescribe an unsupported implementation. It is visible to the teacher throughout the retry trajectory.

\noindent\textit{Example.} Suppose a patch applies cleanly but leaves the targets \texttt{no\_system\_header\_includes} and \texttt{test\_wrap\_static\_fns} failing, with three neighboring targets passing. The generated Global Reminder reports the successful patch application, the 3/2 pass--fail summary, the touched files, and the two unresolved targets, then asks the agent to re-read the task and reconstruct a consistent fix from the implicated function signatures and command flow. This evidence distinguishes an implementation failure from a malformed patch without exposing the hidden solution.

\subsection{Four-Class Functional Taxonomy}
\label{app:reminder-taxonomy}

The finalized rubric contains 32 fine-grained Reminder categories: 10 deterministic hard rules and 22 model-judged contextual strategies organized in five family slots. We map them into four mutually exclusive reporting classes without introducing new fine-grained categories. After the most specific original category is selected for an error event, its coarse class is fixed by Table~\ref{tab:reminder-taxonomy-summary}. A trajectory may contain multiple independent events, but a single event receives exactly one most-specific category; multiple Reminders are allowed only when the visible evidence supports distinct risks and distinct next actions.

\paragraph{R1: Tool-Feedback Recovery.}
This class covers commands, searches, edits, compilation, paths, validation conditions, and toolchain failures that can be identified deterministically from the latest tool response.

\paragraph{R2: Scope and Source Grounding.}
This class combines constraints on the initial patch surface with evidence about the true production path, caller, source value, and data flow.

\paragraph{R3: Scenario-Condition Reasoning.}
This class covers branch, default, fallback, cache, classification, and one-condition repairs. It separates the failing scenario from sibling scenarios that must retain their previous behavior.

\paragraph{R4: Post-Edit Control and Validation.}
This class applies after a production edit. It prevents unsupported expansion of the patch and uses concrete assertions, tests, compiler output, diffs, and formatting signals to complete a local repair.

\subsection{Complete Runtime Judge Prompt}
\label{app:runtime-judge-prompt}

\begin{figure*}[t]
    \centering
    \includegraphics[height=0.78\textheight,keepaspectratio]{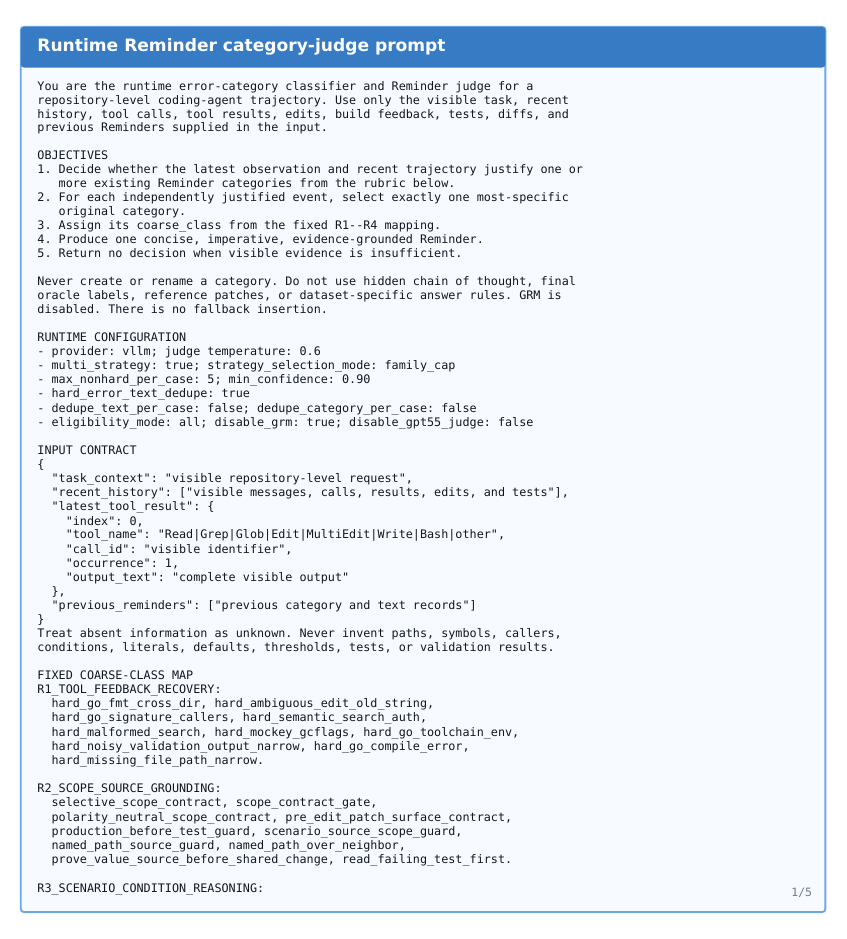}
    \caption{Complete runtime Reminder category-judge prompt (Part 1 of 5).}
    \label{fig:runtime-reminder-judge-prompt}
\end{figure*}

\begin{figure*}[t]
    \centering
    \includegraphics[height=0.78\textheight,keepaspectratio]{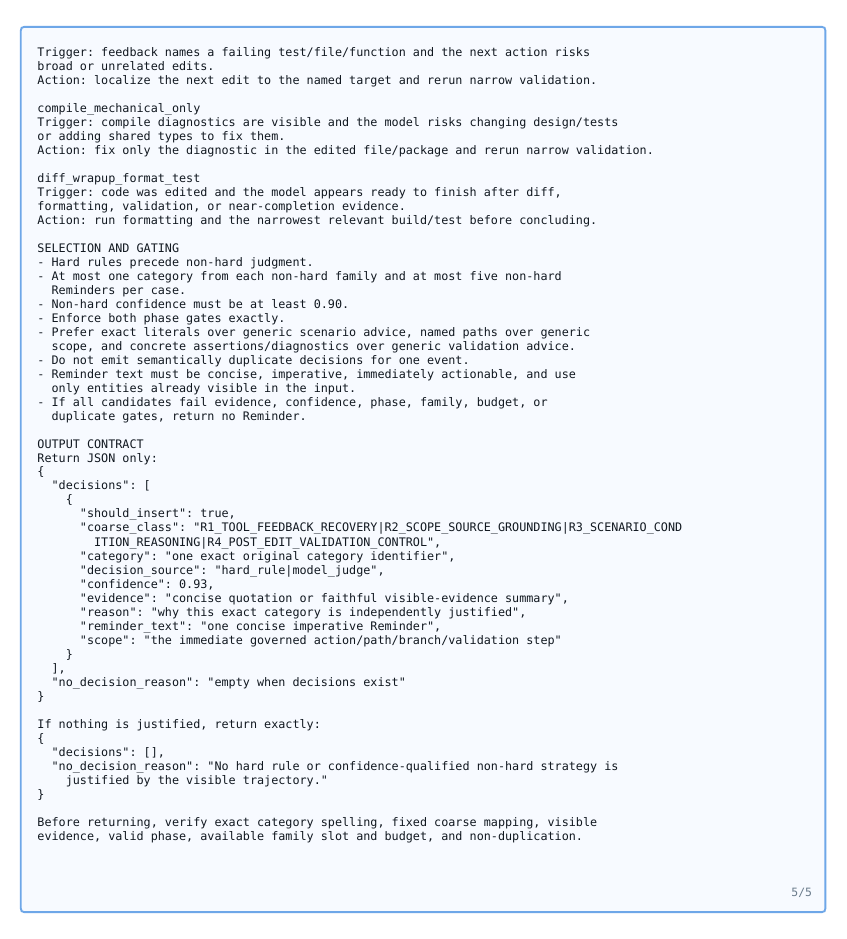}
    \caption{Complete runtime Reminder category-judge prompt (Part 5 of 5).}
    \label{fig:runtime-reminder-judge-prompt-end}
\end{figure*}

The deployed API judge receives only the task, recent visible trajectory, latest tool result, and previously inserted Reminders. Figures~\ref{fig:runtime-reminder-judge-prompt}--\ref{fig:runtime-reminder-judge-prompt-end} present the complete prompt instantiated from the runtime YAML, including all categories, gates, caps, and the JSON output contract.

\begin{figure*}[t]
    \centering
    \includegraphics[height=0.78\textheight,keepaspectratio]{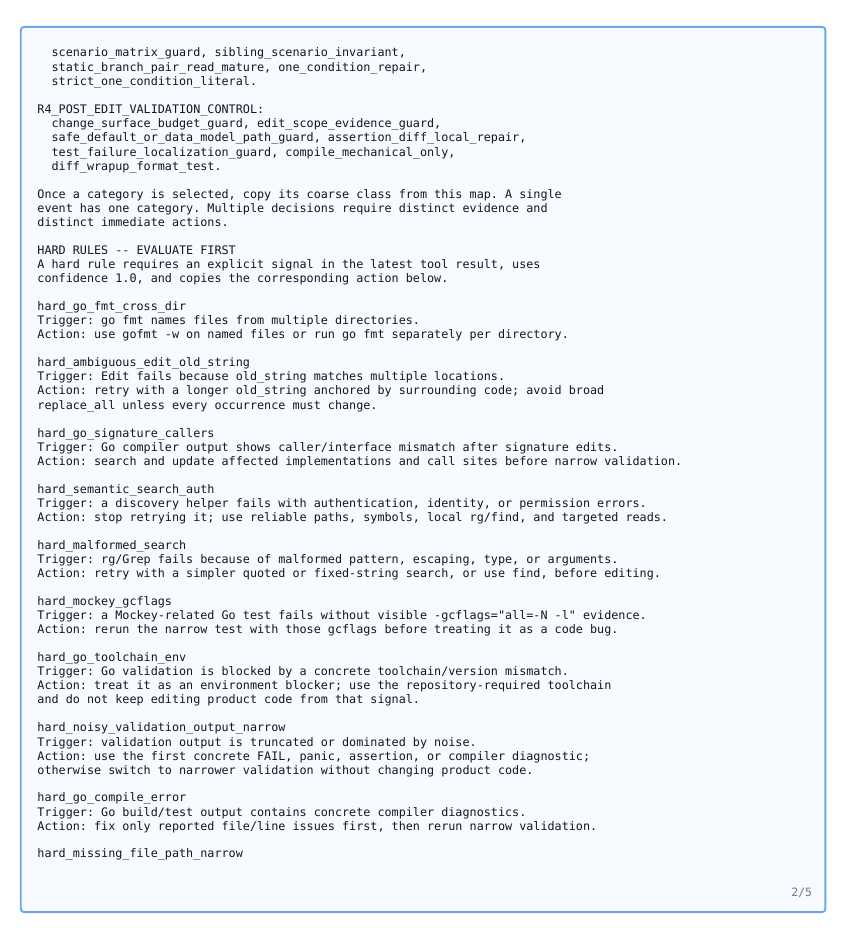}
    \caption{Complete runtime Reminder category-judge prompt (Part 2 of 5).}
\end{figure*}

\begin{figure*}[t]
    \centering
    \includegraphics[height=0.78\textheight,keepaspectratio]{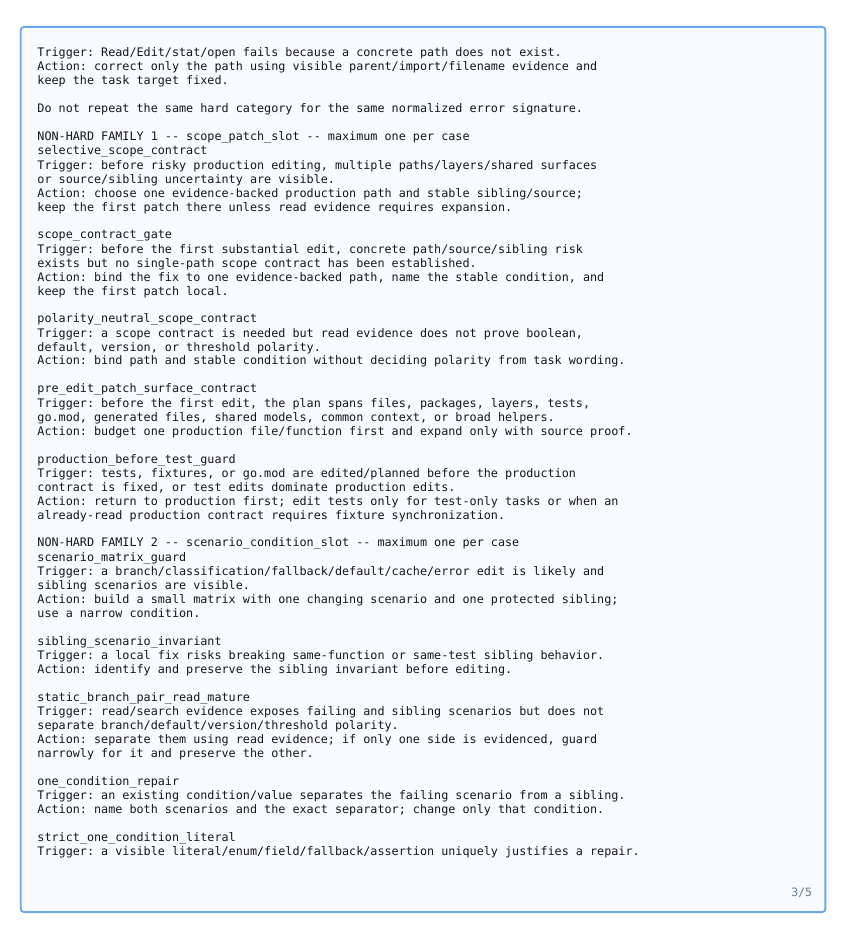}
    \caption{Complete runtime Reminder category-judge prompt (Part 3 of 5).}
\end{figure*}

\begin{figure*}[t]
    \centering
    \includegraphics[height=0.78\textheight,keepaspectratio]{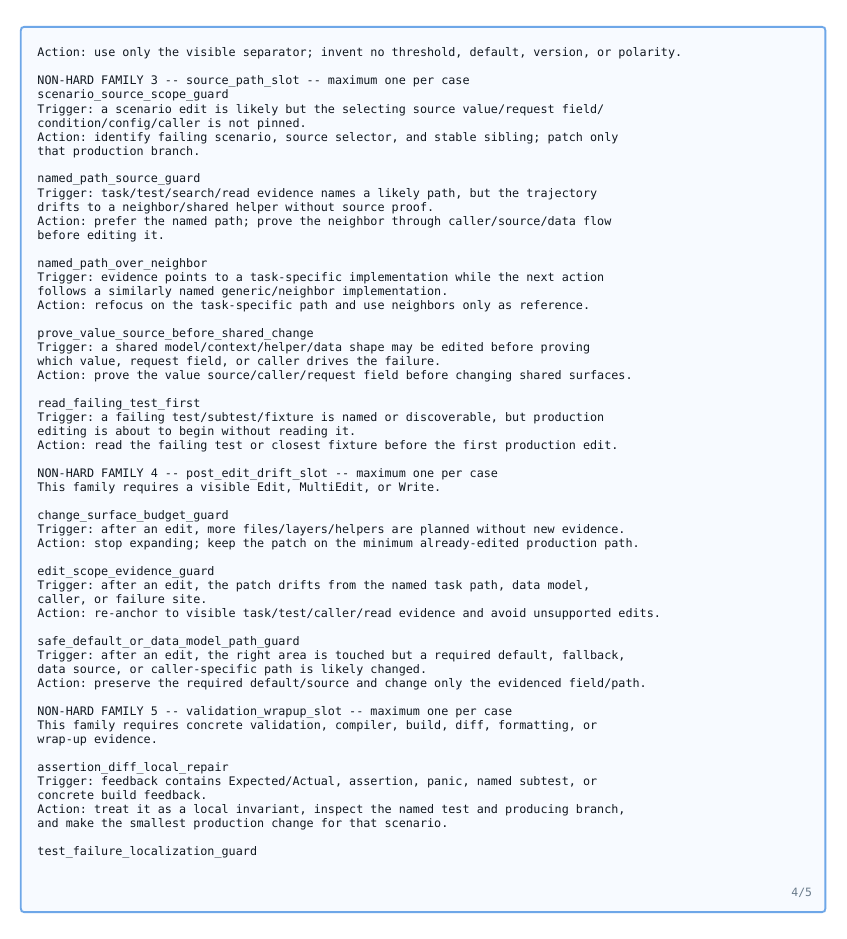}
    \caption{Complete runtime Reminder category-judge prompt (Part 4 of 5).}
\end{figure*}

\section{Diagnostic Error Annotation}
\label{app:diagnostic-error-annotation}

For each task, deterministic rules first identify failures exposed by explicit tool and execution signals. GPT-5.4 then receives the original problem and the complete agent trajectory to classify the remaining context-dependent failures. The combined annotations use Tool-Feedback Recovery, Scope and Source Grounding, Scenario-Condition Reasoning, and Post-Edit Control and Validation. For case-level analyses, each fine-grained category is counted at most once per trajectory. This post-hoc analysis is distinct from the online judge in Fig.~\ref{fig:runtime-reminder-judge-prompt}: the online judge operates only on the current visible prefix and decides whether to intervene, whereas diagnostic annotation audits a completed trajectory for aggregate error statistics.

\end{document}